\documentclass[10pt,twocolumn,letterpaper]{article}

\usepackage[pagenumbers]{cvpr}

\usepackage{cuted}
\usepackage{currfile}
\usepackage{caption}
\usepackage{array}
\usepackage{xspace}

\newcommand{\nscenes}{12}

\newcolumntype{L}[1]{>{\raggedright\arraybackslash}p{#1}}
\newcolumntype{C}[1]{>{\centering\arraybackslash}p{#1}}

\usepackage{microtype}

\definecolor{cvprblue}{rgb}{0.21,0.49,0.74}
\usepackage[pagebackref,breaklinks,colorlinks,allcolors=cvprblue]{hyperref}
\usepackage{multirow}
\usepackage{makecell}

\def\paperID{*****}
\def\confName{CVPR}
\def\confYear{2027}

\newcommand{\method}{RenderBench\xspace}

\newcommand{\best}[1]{\textbf{#1}}

\title{RenderBench: Benchmarking Render-to-Real Video Transfer with Reconstructed Digital Twins}

\author{
Dicong Qiu$^{1,*}$ \quad
Zhiyuan Xu$^{2,*}$ \quad
Yaosheng Liu$^{3}$ \quad
Feng Han$^{4}$ \quad
Bo Ye$^{3}$\\
$^{1}$The Hong Kong University of Science and Technology (Guangzhou) \quad
$^{2}$Jacobi.ai\\
$^{3}$Southeast University \quad
$^{4}$Independent Researcher\\
{\tt\small dqiu570@hkust-gz.edu.cn \quad zhiyuan.xu@jacobi.ai \quad 220266580@seu.edu.cn}\\
{\tt\small Terrancehan25@gmail.com \quad yeb@seu.edu.cn}\\
$^{*}$Equal contribution
}

\begin{document}
\maketitle

\begin{strip}
  \vspace{-8pt}
  \centering
  \includegraphics[width=\textwidth]{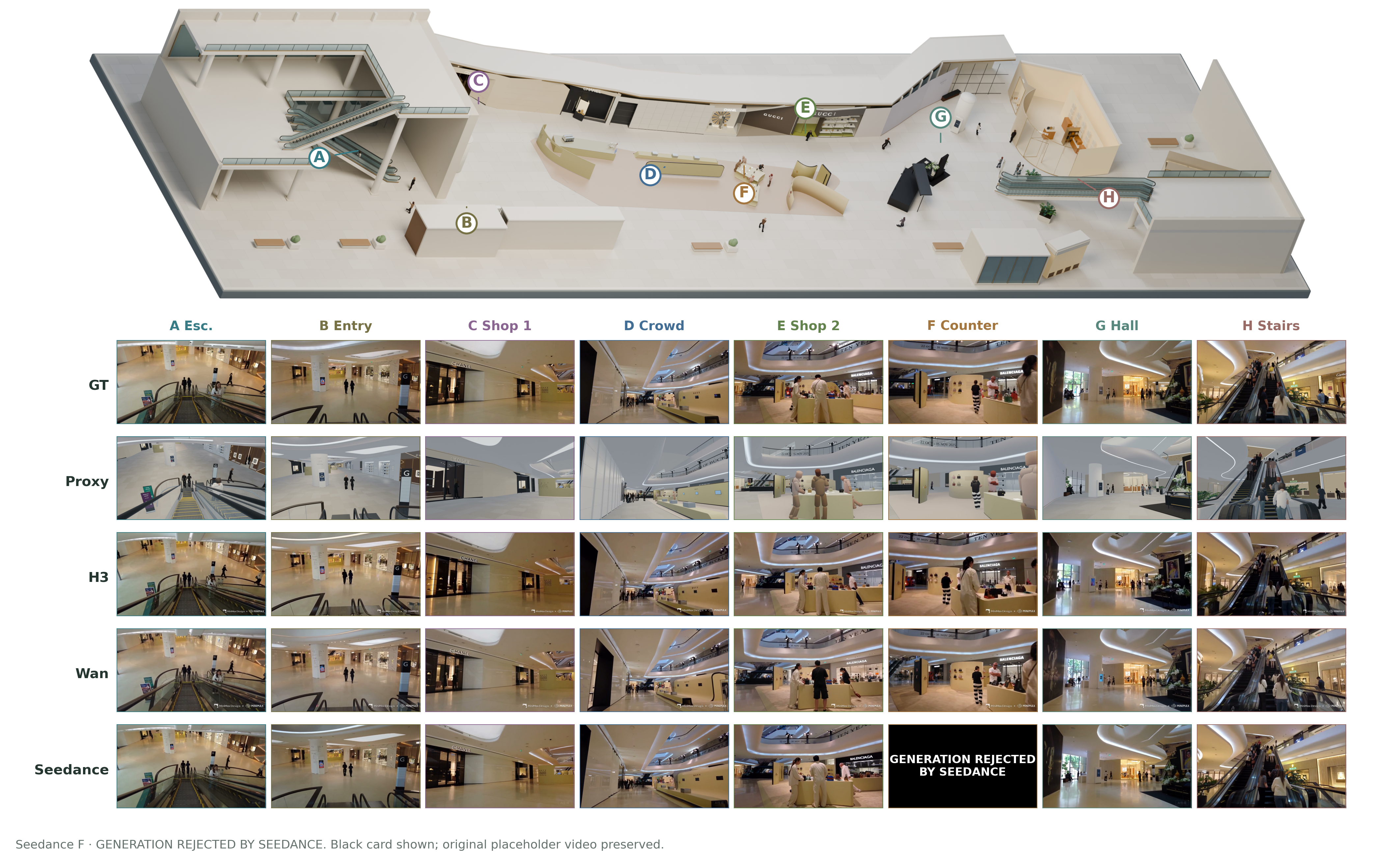}
  \captionsetup{font=small,skip=3pt,hypcap=false}
  \captionof{figure}{Qualitative comparison across eight viewpoints and interactions (A--H) in one reconstructed environment.
  The rows show the real ground truth, aligned proxy render, and outputs from MiniMax H3, Wan2.2-VACE-Fun-A14B, and Seedance 2.5 under matched task prompts.
  The black card marks a rejected Seedance generation.}
  \label{fig:teaser}
  \vspace{-6pt}
\end{strip}

\begin{abstract}
Modern video models can synthesize realistic videos from complementary appearance and geometry conditions: real reference images specify how the scene should look, while proxy RGB and geometric control videos depict its structure, viewpoint changes, and motion.
Yet this reference-conditioned render-to-real capability remains difficult to evaluate because its output must be compared with a ground-truth real video depicting the same scene evolution.
A rigorous benchmark therefore needs appearance references and target real videos paired with editable, geometrically registered 3D replicas, a resource that has traditionally required prohibitive manual modeling, calibration, and animation effort.
Existing evaluations therefore rely on controls estimated from real videos or on small collections of synthetic scenes, neither of which provides paired real appearance and editable scene state at scale.
We introduce \method, a benchmark comprising \nscenes reconstructed real-world scenes that span large-scale indoor environments and egocentric viewpoints, with both dynamic and static settings.
The scenes are constructed using modern visual geometry, neural reconstruction, and assisted 3D authoring.
Each scene is decomposed into static objects and dynamic actors, registered to the capture cameras, and admitted only after multi-view geometric and temporal validation.
An accepted unit contains held-out appearance references, a real target video, an editable digital twin, a matched proxy render, and renderer-native scene annotations.
We evaluate transfer models against paired real target videos, retain PAI-Bench-C-compatible structural projections, and use scene annotations to localize failures by object, visibility, articulation, and motion.
The first release retains 12 of 14 registered samples (85.7\%), comprising 1,496 paired real--proxy frames; all 12 scenes pass file-integrity and environment-edit audits, while proxy diagnostics yield a depth si-RMSE of 0.2170 and instance mIoU of 0.3673.
\end{abstract}

\section{Introduction}
\label{sec:intro}

Computer-graphics simulators offer a compelling source of training data for robotics, autonomous driving, and embodied intelligence.
They provide scalable trajectories together with exact geometry, object state, segmentation, depth, and motion.
Their central limitation is the synthetic-to-real appearance gap~\cite{tobin2017domainrandomization}.
Recent video generators directly target this limitation by combining two complementary sources of information.
One or more real reference images specify the desired appearance, including materials, illumination, and scene style, while a geometry video depicts the scene structure, viewpoint changes, object motion, and articulation to preserve.
The geometry condition may be a proxy RGB render or dense render passes such as depth, segmentation, edges, pose, or trajectories.
Structured-control systems such as Cosmos-Transfer2.5~\cite{nvidia2026cosmostransfer25} and Wan2.2-VACE-Fun-A14B~\cite{alibaba2025wan22fun} expose these signals explicitly, whereas general multimodal systems such as MiniMax H3~\cite{minimax2026h3} and Seedance 2.5~\cite{bytedance2026seedance25} consume them through reference and editing interfaces.
Together, these systems establish reference-conditioned render-to-real transfer as an emerging capability across structured-control and general-purpose video models.

This new capability creates an evaluation problem that existing datasets cannot fully answer.
Existing resources cover complementary pieces of the problem, but not the full evaluation loop.
CO3D provides nearly 19,000 object-centric real videos with recovered cameras and point clouds, but no independently authored, editable proxy whose render can be used as a transfer input~\cite{reizenstein2021co3d}.
Aria Digital Twin (ADT) comes closest to paired real--synthetic supervision: its 200 published sequences include calibrated egocentric recordings, a carefully digitized scene, and photorealistic synthetic renderings~\cite{pan2023adt}.
However, those sequences span only two instrumented spaces, and their construction relies on high-resolution scanning, manual scene digitization, material and lighting tuning, and motion capture; ADT is designed for egocentric 3D perception rather than scalable evaluation from deliberately simplified renderer inputs.
AptSim2Real instead offers approximately paired driving images whose simulated and real scenes may differ in assets, shape, texture, and appearance, making it suitable for distribution-level translation but not same-scene structural verification~\cite{zhang2023aptsim2real}.
Finally, Cosmos TransferBench and PAI-Bench-C evaluate controls extracted from real videos, while the Cosmos Sim2Real case study uses 20 programmatically authored robot scenarios; neither couples those inputs to real captures of the same editable scenes~\cite{nvidia2025cosmostransfer1,zhou2025paibench}.
To determine whether a transfer model merely produces a plausible video or actually reconstructs the intended real-looking scene, evaluation requires four aligned elements: appearance references, a geometry video, a held-out real target video, and an editable 3D scene.
The reference images are model inputs that specify the desired real appearance; the geometry video is the model input that depicts the target viewpoint changes and scene evolution without exposing the underlying camera parameters; the temporally matched real video is held out as the appearance ground truth; and the registered editable 3D scene provides renderer-native geometry and dynamics for diagnostic analysis.
Together they distinguish successful transfer from three common shortcuts: copying the synthetic input without realification, reproducing a reference image without following the requested motion, and generating an attractive but structurally different video.

Historically, this evaluation unit has been impractical to collect at benchmark scale.
Building a faithful, editable 3D replica of an ordinary real scene required skilled artists to model and place assets, reproduce materials and lighting, calibrate cameras, rig articulated agents, and synchronize motion.
Photogrammetry and neural view synthesis reduced parts of this burden, but their outputs were often difficult to edit, animate, or convert into the clean renderer-native state required for controlled evaluation.
The missing resource is therefore not simply a larger video collection, but a repeatable way to build and validate such paired evaluation units across ordinary physical scenes.
Without it, newer structured-control and reference-video models cannot be compared under equivalent scene content with both real appearance evidence and renderer-native structural truth.

The premise of \method is that recent progress changes this feasibility boundary.
3D Gaussian Splatting enables fast high-quality reconstruction and rendering of captured scenes~\cite{kerbl20233dgs}; DUSt3R~\cite{wang2024dust3r} and VGGT~\cite{wang2025vggt} substantially reduce the calibration and geometry burden; and multimodal authoring agents can operate professional 3D tools over long workflows~\cite{openai2026astra}.
These systems do not by themselves guarantee metric fidelity, but they change the labor and iteration cost of producing a candidate digital twin.
Geometric registration and explicit acceptance tests can then convert that candidate into benchmark-grade data.

We therefore make \emph{benchmark construction}, rather than a new scalar metric, the central contribution.
Starting from a captured scene, our assisted pipeline reconstructs editable assets, recovers the camera trajectory, fits object and articulated motion, and iteratively compares rendered views with the observations.
Each scene passes quantitative multi-view and state-consistency gates before release.
The resulting paired data supports appearance comparison to reality and diagnostic analysis against the accepted renderer state.

Our contributions are:
\begin{itemize}
  \item We identify the missing substrate for reference-conditioned render-to-real evaluation: appearance reference images and ground-truth real videos paired with validated, editable, and dynamically registered 3D twins of the same scenes.
  \item We introduce a scalable construction pipeline that combines visual geometry, neural reconstruction, and assisted 3D authoring with explicit geometric and temporal acceptance gates.
  \item We construct \method from \nscenes reconstructed real-world scenes spanning large-scale indoor environments and egocentric viewpoints, with both dynamic and static settings, providing held-out appearance references, paired real target videos and geometry renders, together with renderer-native depth, masks, landmarks, visibility, flow, camera, and articulation metadata.
  \item We benchmark representative transfer systems against paired real target videos, retain PAI-Bench-C-compatible structural projections, and use the paired scene state to analyze where aggregate scores conceal object, articulation, occlusion, and motion failures.
\end{itemize}

\section{Related Work}
\label{sec:related}

\subsection{Render-to-Real Video Generation}

Pix2pix~\cite{isola2017pix2pix}, CycleGAN~\cite{zhu2017cyclegan}, and vid2vid~\cite{wang2018vid2vid} established paired, unpaired, and temporal translation paradigms.
Modern diffusion transfer models substantially improve photorealism, but can still hallucinate geometry or alter motion.
Generic video-generation evaluation includes distributional metrics such as FVD and multi-dimensional suites such as EvalCrafter and VBench~\cite{unterthiner2018fvd,liu2024evalcrafter,huang2024vbench}; unlike \method, they do not assume a frame-aligned real target and editable scene state for each generated video.

Cosmos-Transfer1~\cite{nvidia2025cosmostransfer1} formulates multimodal controllable world generation from visual, edge, depth, and segmentation inputs, explicitly motivating these controls as a way to make CG renders realistic while preserving scene structure and semantics.
Its TransferBench contains 600 real videos and measures control adherence, diversity, and technical quality.
For robotics Sim2Real, it additionally evaluates 20 programmatically generated manipulation scenes with six prompts each.
This work demonstrates both the promise of render-to-real generation and the current construction bottleneck: broad real-video evaluation lacks scene-native 3D truth, while 3D-grounded Sim2Real evaluation remains small and synthetic-only.

Cosmos-Transfer2.5~\cite{nvidia2026cosmostransfer25} extends the specialized Control2World line with multiple spatial video modalities, image prompting, longer autoregressive generation, and simulation-to-photorealism workflows.
Wan2.2-VACE-Fun-A14B~\cite{alibaba2025wan22fun} provides a second open structured-control family with generic control videos, reference images, pose, depth, edge, trajectory, and camera variants.
MiniMax H3~\cite{minimax2026h3} instead exposes a general multimodal reference interface and explicitly supports video-to-video motion transfer.
Seedance 2.5~\cite{bytedance2026seedance25} similarly emphasizes precise video reference, editing, white-model control, and longer generation.
These interfaces are not identical, but all can be asked to realify a proxy scene while retaining specified content or motion.
This diversity strengthens the need for a model-agnostic benchmark and motivates separate structured-control and video-reference tracks.

\subsection{Benchmarks and Paired Real--Synthetic Data}

PAI-Bench~\cite{zhou2025paibench} generalizes this evaluation into generation, conditional-generation, and understanding tracks for Physical AI.
PAI-Bench-C extracts blur, edge, depth, and segmentation controls from real videos and reports Blur SSIM, Edge F1, Depth si-RMSE, Mask mIoU, DOVER technical quality, and LPIPS diversity.
\method retains the edge, depth, and segmentation projections for compatibility, but uses paired real target videos as its primary reference rather than adopting the complete protocol or proposing a new aggregate metric.
Its contribution is a different evaluation substrate: paired real observations and editable 3D twins that allow errors to be analyzed by object, visibility, articulation, and motion.

CO3D~\cite{reizenstein2021co3d} supplies large-scale object-centric real videos, camera poses, and reconstructed point clouds for category-level reconstruction and novel-view synthesis, but does not standardize an independently built proxy-render input.
ADT~\cite{pan2023adt} supplies unusually accurate real--synthetic correspondence, full calibration, scene and object geometry, and photorealistic renderings.
It is the closest data precedent to ours, but differs in both objective and construction regime: ADT targets egocentric perception in two intensively digitized, motion-captured spaces, whereas \method targets scalable diagnosis of transfer from deliberately simplified, editable renders of diverse captured scenes.
AptSim2Real~\cite{zhang2023aptsim2real} studies approximately paired image translation; its real and simulated examples need not depict the identical geometry, precluding renderer-native measurement of same-scene preservation.

\subsection{Building Editable Digital Twins}

Classical structure from motion and multi-view stereo provide geometric scene reconstruction but require sufficiently overlapping imagery and substantial cleanup~\cite{schoenberger2016sfm}.
3D Gaussian Splatting~\cite{kerbl20233dgs} improves the quality and speed of novel-view reconstruction, and Dynamic 3D Gaussians extends this representation to persistent dynamic tracking~\cite{luiten2024dynamic3dgaussians}, while DUSt3R~\cite{wang2024dust3r} directly predicts point maps from uncalibrated image collections.
VGGT~\cite{wang2025vggt} further predicts cameras, depth, point maps, and tracks in a feed-forward pass.
These representations supply geometry and appearance evidence, but benchmark construction also needs editable objects, semantic organization, clean topology, articulation, and reproducible rendering.

GPT-6 Astra is a general reasoning and computer-use model rather than a dedicated geometric estimator.
Nevertheless, its demonstrated ability to reconstruct CAD objects from multi-view renders and carry out Blender-to-Unreal scene workflows~\cite{openai2026astra} is important for our setting: an agent can connect specialized reconstruction outputs with asset modeling, scene organization, scripting, rendering, and visual iteration.
\method treats this capability as one possible construction accelerator, not as ground truth or a required model component; every resulting twin must pass the same independent validation.

\begin{figure*}[t]
  \centering
  \includegraphics[width=\textwidth]{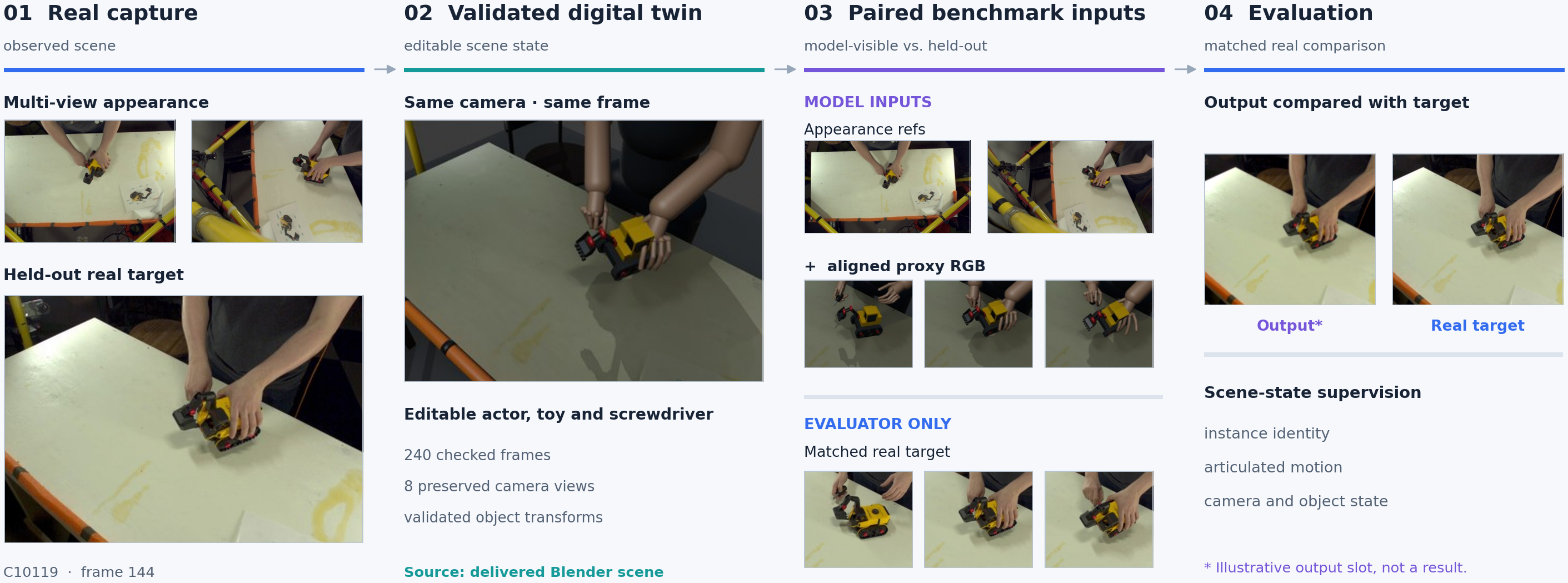}
  \caption{Construction and evaluation pipeline of \method.
  Multi-view observations are converted into a validated digital twin with editable objects, articulated actors, and aligned cameras.
  Transfer models receive appearance references and an aligned proxy video, while the matched real target remains evaluator-only.
  Generated outputs are compared directly with the real target using paired-reference metrics, while PAI-Bench-C-compatible structural projections and paired scene state support diagnostic analysis.}
  \label{fig:pipeline}
\end{figure*}
\section{Benchmark Setting}
\label{sec:problem}

\subsection{Paired Real--3D Scene Unit}
Each benchmark unit contains appearance reference images $I_{1:K}^{\mathrm{ref}}$, a ground-truth real target video $Y_{1:T}^{\mathrm{gt}}$, and a reconstructed dynamic digital twin $\mathcal S_{1:T}$ registered to the target video.
The reference images depict the target scene appearance but are captured from frames or viewpoints excluded from the scored target sequence.
The recovered camera trajectory $C_{1:T}$ is applied in a proxy renderer $R_p$ to produce the geometry-proxy video:
\begin{equation}
 X_{1:T}^{\mathrm{geo}}=R_p(\mathcal S_{1:T},C_{1:T}).
\end{equation}
Depending on the evaluated interface, $X^{\mathrm{geo}}$ may contain proxy RGB alone or a fixed bundle of rendered geometry channels such as depth, segmentation, edges, pose, or object trajectories.
An evaluated transfer system produces
\begin{equation}
 \hat Y_{1:T}=G(I_{1:K}^{\mathrm{ref}},X_{1:T}^{\mathrm{geo}}).
\end{equation}
Camera intrinsics, extrinsics, and trajectories are never exposed to the transfer model; it must infer viewpoint change from $X^{\mathrm{geo}}$ itself.

The desired output follows the appearance specified by $I^{\mathrm{ref}}$, matches the held-out real target $Y^{\mathrm{gt}}$, and preserves the scene and trajectory encoded by $(\mathcal S,C)$.
Because reconstruction is never perfect, we publish a validity mask and registration uncertainty for every sequence; evaluation excludes pixels and states outside the accepted support.

\subsection{Renderer-Native Scene Annotations}
Separately from the model inputs, the benchmark retains
\begin{equation}
 A_t^{\mathrm{eval}}=\{D_t,M_t,P_t,V_t,F_{t\rightarrow t+1},Q_t,C_t\},
\end{equation}
denoting canonical depth, instance masks, projected landmarks, visibility, optical flow, object/articulation state, and camera parameters.
These annotations are available only for evaluation and analysis.
If a visualized depth, mask, or pose channel is included in $X^{\mathrm{geo}}$, it is treated as part of the declared geometry-proxy representation; $A^{\mathrm{eval}}$ remains evaluator-only reference information rather than an additional input to $G$.
We use these annotations to stratify and interpret paired-reference measurements, including PAI-Bench-C-compatible structural projections, rather than to define a new composite score~\cite{zhou2025paibench}.

\subsection{Conditioning and Evaluation Roles}
The appearance images, real target video, and digital twin serve distinct roles.
The model observes $I^{\mathrm{ref}}$ as the appearance condition but never observes the scored frames of $Y^{\mathrm{gt}}$.
The ground-truth video evaluates illumination, material response, sensor appearance, image detail, and temporal realism along the requested trajectory.
The accepted 3D scene provides a reference for analyzing geometry, object identity, visibility, motion, and articulation.
No source is silently treated as perfect outside its designated role.

\subsection{Task Setting}
At test time, every transfer model receives exactly two semantic inputs: the real appearance references $I^{\mathrm{ref}}$ and the geometry-proxy video $X^{\mathrm{geo}}$.
The model must infer the target viewpoint change and scene evolution from $X^{\mathrm{geo}}$, render them in the appearance specified by $I^{\mathrm{ref}}$, and produce $\hat Y$ without observing $Y^{\mathrm{gt}}$, $A^{\mathrm{eval}}$, or explicit camera information.
The benchmark fixes the number of reference images and the representation of $X^{\mathrm{geo}}$ for each comparison.
Different model APIs may package the pair through image-reference, video-editing, or structured-control interfaces, but adapters may not introduce any additional semantic condition.

\section{Constructing RenderBench}
\label{sec:benchmark}

\subsection{Why Construction Was the Bottleneck}
The desired benchmark is not a collection of attractive 3D assets.
Every scene must be recognizably and geometrically tied to a real capture, editable enough to expose semantic objects and articulation, and accurate enough to render the observed camera path.
Traditional manual construction satisfies these requirements but does not scale at an acceptable cost.
Pure neural view synthesis scales better but generally lacks the clean meshes, semantics, joints, and physical editability needed for controlled rendering.

\subsection{Digital-Twin Construction Pipeline}

Given synchronized multi-view observations, the pipeline constructs an editable digital twin containing static geometry, semantic object instances, materials, lighting, calibrated cameras, and time-varying object and articulation states.
Static geometry and dynamic actors are reconstructed separately and then registered in a common metric coordinate system.
This separation is important: static objects require accurate shape and layout, whereas dynamic actors additionally require temporally stable identity and articulation.

\paragraph{Capture and initialization.}
We record calibrated multi-view images or videos with scale anchors and synchronized views for dynamic sequences where available.
DUSt3R/VGGT initialize cameras, geometry, depth, and tracks~\cite{wang2024dust3r,wang2025vggt}, while 3D Gaussian Splatting provides an appearance reference for novel-view comparison~\cite{kerbl20233dgs}.
These estimates are initializations rather than benchmark ground truth.
Camera parameters and scene scale are jointly checked against multi-view correspondences and physical anchors before downstream reconstruction.

\paragraph{Static scene reconstruction.}
The scene is decomposed into architecture and semantic object instances.
Multi-view masks and reconstruction estimates initialize each object's geometry and pose.
The geometry and layout are refined against silhouettes, robust depth agreement, and multi-view correspondences, with regularization used where observations are incomplete.
The optimization is performed per object and jointly for scene layout, so a low whole-image error cannot hide a misplaced small object.
Stable instance identities are retained across views, scene assets, and renderer annotations.

\begin{table*}[t]
\centering
\caption{Dataset release audit and proxy-quality diagnostics. Bracketed values denote the reported scene-level intervals.}
\label{tab:construction}
\small
\begin{tabular}{@{}L{0.22\textwidth}L{0.51\textwidth}r@{}}
\toprule
Category & Audit item & Result\\
\midrule
Dataset scale & Registered samples / released scenes & 14 / 12\\
& Sample retention rate & 85.7\%\\
& Paired real--proxy frames & 1,496\\
& Video frame rate & 30 fps\\
\addlinespace
File verification & Real and proxy frame counts match & 12 / 12 scenes\\
& File SHA-256 matches the release manifest & 48 / 48 files\\
\addlinespace
Environment audit & Background-ray hit rate & 100.0\% (502,656 rays)\\
& Measurable background shadow response & 12 / 12 scenes\\
& Environment visual inspection passed & 12 / 12 scenes (36 frames)\\
& Foreground, camera, and animation hashes unchanged after environment edits & 12 / 12 scenes\\
\addlinespace
Proxy diagnostics & Estimated depth si-RMSE $\downarrow$ & 0.2170 [0.1774, 0.2633]\\
& Estimated instance mIoU $\uparrow$ & 0.3673 [0.2751, 0.4504]\\
& Estimated instance coverage $\uparrow$ & 0.6228 [0.4862, 0.7426]\\
\bottomrule
\end{tabular}
\end{table*}

\paragraph{Dynamic actor reconstruction.}
Humans and robots are reconstructed as sequence-level articulated models rather than independent per-frame meshes.
A fixed topology and articulation structure are initialized once, after which root pose and joint states $\mathcal Q_{1:T}$ are fitted to multi-view keypoints, silhouettes, and trajectories.
Temporal regularization suppresses implausible velocity and acceleration, while contact constraints align grasp, release, and support events with manipulated objects.
Visibility weights suppress unsupported constraints during occlusion instead of forcing a per-frame fit.
Fixed topology, link lengths, and semantic attachments prevent identity and scale drift while preserving editable articulation.

\paragraph{Agent-guided refinement.}
The construction agent proposes candidate scene edits from reconstruction residuals and diagnostic renders, while ambiguous identities and topology are resolved by an operator.
The same acceptance tests apply regardless of how a candidate was produced.
Geometry and motion are frozen before materials and lighting are refined, preventing appearance optimization from concealing structural errors.

\paragraph{Registration and acceptance.}
Static geometry, articulated motion, cameras, materials, and lighting are assembled into an editable digital twin.
Validation uses views and frames withheld from fitting whenever capture coverage permits.
A scene is accepted only when camera reprojection, masked depth agreement, instance silhouette overlap, landmark accuracy, temporal smoothness, event timing, and held-out-view appearance satisfy pre-registered thresholds.
Acceptance is evaluated both globally and per semantic object, with visibility-dependent thresholds for small or partially occluded instances.
Regions that remain ambiguous are excluded through the released validity masks rather than silently treated as accurate geometry.
Failed scenes are repaired or excluded, and construction error is propagated to the uncertainty analysis in Sec.~\ref{sec:metrics}.

\subsection{Dataset Contents}
\begin{figure*}[t]
  \centering
  \includegraphics[width=\textwidth]{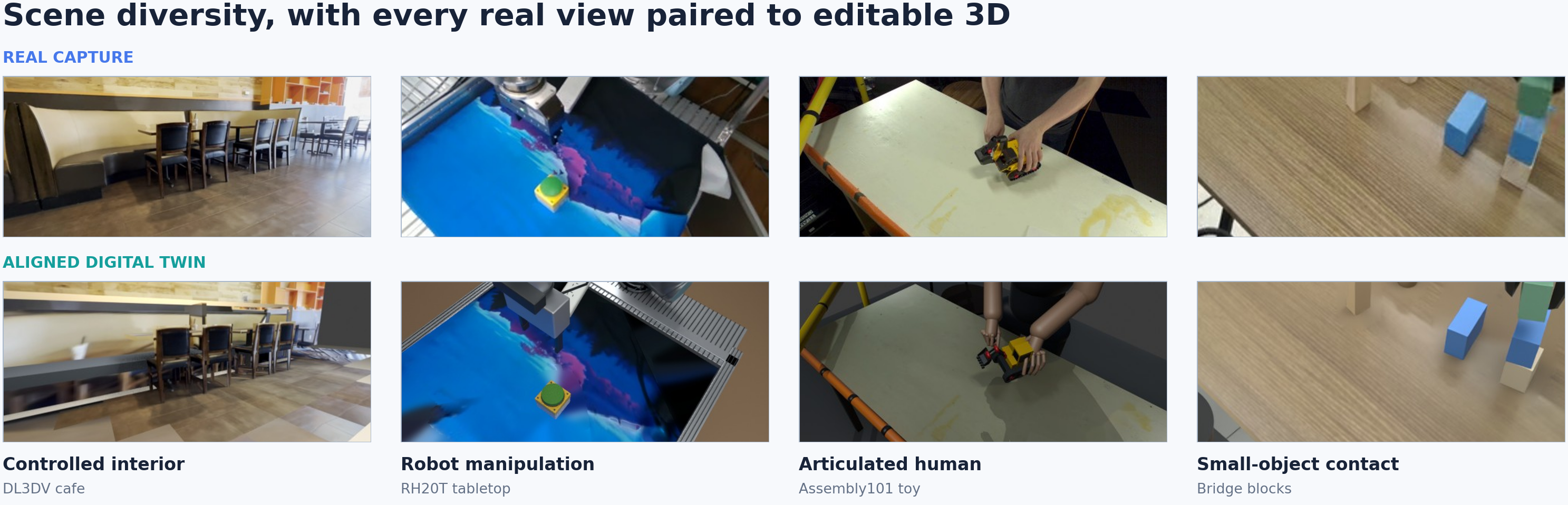}
  \caption{Dataset diversity and paired supervision in \method.
  The \nscenes reconstructed scenes span large-scale indoor environments and egocentric viewpoints, with both dynamic and static settings.
  Each real capture is paired with an editable digital twin, aligned proxy renders, and renderer-native scene annotations.}
  \label{fig:dataset-overview}
\end{figure*}

Each accepted unit contains held-out appearance reference images, the ground-truth real target video, an editable scene package, calibrated camera trajectory, and the proxy RGB geometry video.
Its evaluator-only annotations include depth, normals, optical flow, semantic/instance masks, stable object identifiers, landmarks, visibility, articulation state, event labels, validity masks, and uncertainty estimates.
Construction metadata records capture provenance, coordinate conventions, and validation results.
The first release contains \nscenes reconstructed real-world scenes spanning large-scale indoor environments and egocentric viewpoints, including dynamic scenes with moving or articulated actors as well as static scenes.
Figure~\ref{fig:dataset-overview} summarizes the dataset diversity and paired supervision.

\subsection{Model-Interface Adapters}
The same scene unit is packaged through standardized adapters rather than forcing every model into one API.
For structured-control models, adapters provide a fixed budget of real appearance reference images and export the supported proxy RGB, depth, segmentation, edge, pose, and object-trajectory videos with fixed normalization.
For general reference models, adapters provide the same appearance references and the proxy RGB geometry clip.
Adapters never expose camera intrinsics, extrinsics, or trajectories; viewpoint change is encoded only through the rendered geometry videos.
Every submission manifest records the information actually exposed to the model.
Primary comparisons are made within matched information budgets; cross-track plots are descriptive rather than leaderboard ranks.

\subsection{Cost and Scalability Audit}
For every scene we log wall-clock construction time, human interaction time, model/API and GPU cost, number of agent iterations, repair count, and acceptance outcome.
We compare the full pipeline against manual-from-scratch authoring and reconstruction without agentic assistance on a representative subset.
The main construction result reports cost per accepted scene, not cost per attempted scene.

\subsection{Splits and Leakage}
Splits are disjoint in physical location, captured trajectory, asset identity, and human/robot motion.
Appearance references are separated from scored target frames by viewpoint or time while preserving the scene's material and illumination setting.
For dynamic scenes, reference images exclude the scored motion interval, and the exact target camera path is never included among the references.
We additionally reserve a held-out proxy renderer.
If public assets or pretrained reconstruction models may have seen the capture location, this is disclosed; exact benchmark captures and camera paths remain held out.

\begin{table*}[t]
\centering
\caption{Paired-reference results against the aligned real target videos. Seedance 2.5 is evaluated using its $(0,0)$-aligned output.}
\label{tab:main}
\small
\resizebox{\textwidth}{!}{%
\begin{tabular}{lccccccc}
\toprule
Model & LPIPS-GT$\downarrow$ & MS-SSIM-GT$\uparrow$ & Lab $\Delta E_{2000}$$\downarrow$ & Edge F1$\uparrow$ & Depth si-RMSE$\downarrow$ & Inst. mIoU$\uparrow$ & Track err.$\downarrow$\\
\midrule
Proxy render & 0.2683 & 0.7607 & 10.8193 & 0.1987 & 0.2170 & 0.3673 & 0.0089\\
Cosmos-Transfer2.5 & 0.2728 & 0.7868 & 9.4432 & 0.1529 & 0.2532 & 0.3674 & 0.0077\\
Wan2.2-VACE-Fun-A14B & 0.1897 & 0.8612 & 5.5227 & 0.2482 & 0.1566 & 0.4477 & 0.0090\\
MiniMax H3 & 0.1229 & 0.8996 & 4.6114 & 0.4556 & \best{0.1142} & \best{0.6202} & \best{0.0068}\\
Seedance 2.5 & \best{0.1188} & \best{0.9049} & \best{2.7857} & \best{0.4810} & 0.1334 & 0.6142 & 0.0073\\
\bottomrule
\end{tabular}%
}
\end{table*}
\section{Evaluation Protocol}
\label{sec:metrics}

\subsection{Paired Real-Video Reference}
Every generated video $\hat Y_{1:T}$ is evaluated against its frame-aligned real target $Y^{\mathrm{gt}}_{1:T}$.
All comparisons use the released validity masks, and the same preprocessing or estimator is applied to the generated and real videos.
The renderer annotations are used to define valid support and diagnostic subsets, not as a replacement for the real-video reference.

\subsection{Appearance Fidelity}
We measure perceptual similarity using masked LPIPS~\cite{zhang2018lpips} and report lower values as better.
We additionally compute masked MS-SSIM-GT between each generated frame and its aligned real target~\cite{wang2003msssim}, together with CIEDE2000 color error~\cite{sharma2005ciede2000}.
MS-SSIM-GT measures luminance, contrast, and structural agreement across multiple spatial scales, while LPIPS captures perceptual discrepancy and Lab error isolates color mismatch.

\subsection{Structural Fidelity}
Following the edge, depth, and segmentation dimensions of PAI-Bench-C~\cite{zhou2025paibench}, we apply identical projections to $\hat Y$ and $Y^{\mathrm{gt}}$ and report tolerant Edge F1, estimated-depth si-RMSE, and Instance mIoU.
Edge matching uses a small spatial tolerance to avoid penalizing sub-pixel registration error.
Depth is compared after the scale convention used by PAI-Bench-C, and masks are aggregated per instance so that large backgrounds cannot hide missing small objects.
We additionally report instance detection coverage for objects that cannot be matched in the generated video; coverage is a diagnostic quantity reported alongside, but not merged into, the main table.

\subsection{Motion Fidelity}
The same tracker and landmark detector are applied to the generated and real videos.
We report object-normalized trajectory error and track coverage for rigid objects, articulated joints, hands, and robot end effectors.
These are paired-reference measurements: no reference-free temporal quality or warping score is used.

\subsection{Aggregation and Diagnostic Slices}
Metrics are aggregated over valid frames within a run, then over seeds and scenes; confidence intervals are obtained by bootstrap resampling scenes.
Primary values are accompanied by slices over foreground and background, object size, visibility, rigid and articulated parts, and low- and high-motion intervals.
We retain the individual measurements and do not combine them into a single score.

\section{Experiments}
\label{sec:experiments}

\subsection{Research Questions}
\textbf{Q1: Feasibility.} Can the proposed workflow produce accepted real-scene digital twins at substantially lower human time and total cost than conventional authoring?
\textbf{Q2: Benchmark value.} Do paired real-video and scene annotations reveal where similar aggregate scores arise from different transfer failures?
\textbf{Q3: Model capability.} Which transfer models best reproduce the referenced real appearance while preserving the geometry and dynamics of the input video?
\textbf{Q4: Generalization.} Do conclusions persist across unseen scenes, motions, and proxy renderers?

\subsection{Digital-Twin Construction Evaluation}
Table~\ref{tab:construction} reports the release scale, file-integrity checks, environment-edit audits, and proxy-quality diagnostics.
Of 14 registered samples, 12 are released, yielding an 85.7\% retention rate and 1,496 paired real--proxy frames at 30 fps.
All released scenes pass frame-count, manifest-hash, background-coverage, shadow-response, visual-inspection, and edit-isolation checks.
We additionally report scene-level intervals for estimated depth and instance agreement to expose variation hidden by aggregate proxy scores.

\subsection{Transfer Evaluation}
\begin{figure*}[t]
  \centering
  \includegraphics[width=\textwidth]{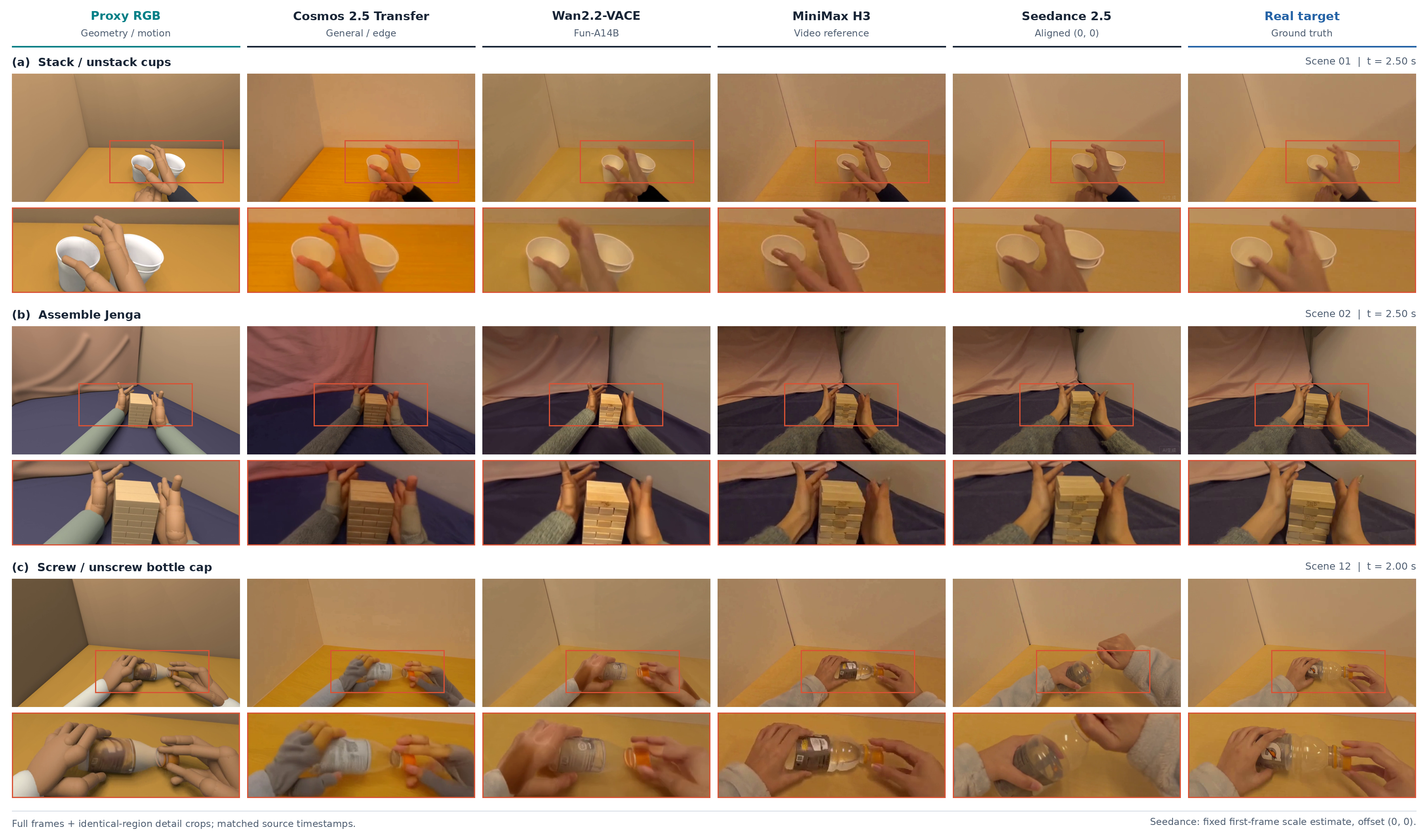}
  \caption{Qualitative comparison on representative paired scenes.
  The proxy input specifies geometry and motion, while the held-out real target provides the ground-truth video for evaluation.
  Zoomed regions highlight structural, articulation, and appearance failures that may be obscured by aggregate scores.}
  \label{fig:qualitative-results}
\end{figure*}

We evaluate representative structured-control and general video-reference systems: Cosmos-Transfer2.5~\cite{nvidia2026cosmostransfer25}, Wan2.2-VACE-Fun-A14B~\cite{alibaba2025wan22fun}, MiniMax H3~\cite{minimax2026h3}, and Seedance 2.5~\cite{bytedance2026seedance25}.
The first two populate the structured-control track; H3 and Seedance populate the video-reference/editing track; any model supporting multiple interfaces is evaluated in each applicable track.
Within each track, models receive identical appearance-reference budgets, source geometry renders, duration, resolution, and geometry-channel budgets; no model receives explicit camera information, and stochastic models use multiple runs.
Primary results are reported in Table~\ref{tab:main}.
Figure~\ref{fig:qualitative-results} provides a qualitative comparison on representative scenes.

\subsection{Paired-Scene Diagnostic Analysis}
We use accepted scene annotations to partition paired-reference errors by object size, visibility, articulation, and motion.
Controlled perturbations to camera pose, object placement, joint angle, blur, texture, and detail test whether these diagnostic slices respond to the intended failure.
We report sensitivity to perturbation severity and agreement with the corresponding appearance, structural, or trajectory metric rather than introducing an additional ranking score.

\subsection{Generalization and Human Validation}
Results are stratified by construction error, motion, occlusion, articulation, reflective/transparent materials, and held-out renderer.
Human raters answer separate questions about real-scene resemblance, structural faithfulness, fine detail, and temporal artifacts.
We correlate each judgment with its corresponding metric and report scene-bootstrap confidence intervals.

\section{Discussion and Limitations}
\label{sec:discussion}

\paragraph{Assisted authoring changes cost, not the reference.}
Agentic 3D authoring can accelerate scene construction but can still choose an incorrect asset, conceal a discrepancy with appearance, or create invalid topology.
It therefore produces candidates, while calibrated observations and acceptance tests determine benchmark inclusion.

\paragraph{Digital-twin error.}
No reconstructed scene is identical to reality.
We publish uncertainty and validity support, exclude unresolved regions, and analyze how benchmark conclusions change across stricter acceptance thresholds.
Claims concern fidelity to an accepted twin and its registered capture, not complete recovery of hidden physical state.

\paragraph{Appearance alignment.}
Even with a matched camera path, exposure, rolling shutter, motion blur, transparent surfaces, and non-Lambertian effects complicate pixel alignment.
Multi-scale metrics and validity masks reduce but do not eliminate this problem.

\paragraph{Scalability.}
The construction audit reports failure rate, human time, total cost, and comparison with manual authoring, since successful examples alone do not establish a scalable benchmark pipeline.

\paragraph{Benchmark scope.}
The first release focuses on scenes that can pass geometric and temporal validation.
Fluids, deformable objects, outdoor crowds, severe occlusion, and uninstrumented fast motion remain difficult.
We expose this selection bias in the dataset statistics and per-category results.

\paragraph{Heterogeneous model interfaces.}
Structured controls and general video references expose different amounts and kinds of information.
A single pooled ranking would confound model quality with conditioning bandwidth.
Track-specific reporting and matched information budgets improve fairness, but cannot make closed natural-language interfaces exactly equivalent to explicit geometry controls.

\paragraph{Licensing and privacy.}
Real captures, reconstructed locations, 3D assets, and human motion require explicit permissions and compatible release terms.
We remove identifying biometric and location information where consent does not cover public release.

\section{Conclusion}
\label{sec:conclusion}

Render-to-real generation has advanced faster than the data needed to evaluate it.
We introduced \method around the missing substrate: held-out real appearance references and ground-truth target videos paired with validated, editable, and dynamically registered 3D twins.
Recent visual-geometry, neural-reconstruction, and assisted authoring systems reduce the cost of producing candidate scenes.
Independent geometric gates then provide distinct appearance conditions, paired real-video targets, and renderer-native scene annotations.
We evaluate transfer quality against paired real target videos, retain PAI-Bench-C-compatible structural projections, and use scene annotations to localize errors across objects, visibility, articulation, and motion.
The audited release contains 12 accepted scenes and 1,496 paired real--proxy frames, with complete frame-count and SHA-256 verification across all released files.
Environment edits preserve foreground, camera, and animation hashes in every scene, and the measured proxy diagnostics establish a transparent baseline for interpreting downstream transfer results.

{\small
\bibliographystyle{bst/ieeenat_fullname}
\bibliography{references}

@inproceedings{wang2003msssim,
  author    = {Zhou Wang and Eero P. Simoncelli and Alan C. Bovik},
  title     = {Multiscale Structural Similarity for Image Quality Assessment},
  booktitle = {Asilomar Conference on Signals, Systems and Computers},
  year      = {2003},
  volume    = {2},
  pages     = {1398--1402}
}

@inproceedings{schoenberger2016sfm,
  author    = {Johannes L. Sch{\"o}nberger and Jan-Michael Frahm},
  title     = {Structure-from-Motion Revisited},
  booktitle = {CVPR},
  year      = {2016}
}

@inproceedings{luiten2024dynamic3dgaussians,
  author    = {Jonathon Luiten and Georgios Kopanas and Bastian Leibe and Deva Ramanan},
  title     = {Dynamic 3D Gaussians: Tracking by Persistent Dynamic View Synthesis},
  booktitle = {3DV},
  year      = {2024}
}

@article{sharma2005ciede2000,
  author  = {Gaurav Sharma and Wencheng Wu and Edul N. Dalal},
  title   = {The {CIEDE2000} Color-Difference Formula: Implementation Notes, Supplementary Test Data, and Mathematical Observations},
  journal = {Color Research \& Application},
  year    = {2005},
  volume  = {30},
  number  = {1},
  pages   = {21--30}
}

@inproceedings{zhang2018lpips,
  author    = {Richard Zhang and Phillip Isola and Alexei A. Efros and Eli Shechtman and Oliver Wang},
  title     = {The Unreasonable Effectiveness of Deep Features as a Perceptual Metric},
  booktitle = {CVPR},
  year      = {2018}
}

@article{unterthiner2018fvd,
  author  = {Thomas Unterthiner and Sjoerd van Steenkiste and Karol Kurach and Rapha{\"e}l Marinier and Marcin Michalski and Sylvain Gelly},
  title   = {Towards Accurate Generative Models of Video: A New Metric and Challenges},
  journal = {arXiv preprint arXiv:1812.01717},
  year    = {2018}
}

@inproceedings{huang2024vbench,
  author    = {Ziqi Huang and Yinan He and Jiashuo Yu and Fan Zhang and Chenyang Si and Yuming Jiang and Yuanhan Zhang and Tianxing Wu and Qingyang Jin and Nattapol Chanpaisit and Yaohui Wang and Xinyuan Chen and Limin Wang and Dahua Lin and Yu Qiao and Ziwei Liu},
  title     = {{VBench}: Comprehensive Benchmark Suite for Video Generative Models},
  booktitle = {CVPR},
  year      = {2024}
}

@inproceedings{liu2024evalcrafter,
  author    = {Yaofang Liu and Xiaodong Cun and Xuebo Liu and Xintao Wang and Yong Zhang and Haoxin Chen and Yang Liu and Tieyong Zeng and Raymond Chan and Ying Shan},
  title     = {{EvalCrafter}: Benchmarking and Evaluating Large Video Generation Models},
  booktitle = {CVPR},
  year      = {2024}
}

@inproceedings{isola2017pix2pix,
  author    = {Phillip Isola and Jun-Yan Zhu and Tinghui Zhou and Alexei A. Efros},
  title     = {Image-to-Image Translation with Conditional Adversarial Networks},
  booktitle = {CVPR},
  year      = {2017}
}

@inproceedings{zhu2017cyclegan,
  author    = {Jun-Yan Zhu and Taesung Park and Phillip Isola and Alexei A. Efros},
  title     = {Unpaired Image-to-Image Translation Using Cycle-Consistent Adversarial Networks},
  booktitle = {ICCV},
  year      = {2017}
}

@inproceedings{wang2018vid2vid,
  author    = {Ting-Chun Wang and Ming-Yu Liu and Jun-Yan Zhu and Guilin Liu and Andrew Tao and Jan Kautz and Bryan Catanzaro},
  title     = {Video-to-Video Synthesis},
  booktitle = {NeurIPS},
  year      = {2018}
}

@inproceedings{reizenstein2021co3d,
  author    = {Jeremy Reizenstein and Roman Shapovalov and Philipp Henzler and Luca Sbordone and Patrick Labatut and David Novotny},
  title     = {Common Objects in 3D: Large-Scale Learning and Evaluation of Real-Life 3D Category Reconstruction},
  booktitle = {ICCV},
  pages     = {10901--10911},
  year      = {2021}
}

@inproceedings{pan2023adt,
  author    = {Xiaqing Pan and Nicholas Charron and Yongqian Yang and Scott Peters and Thomas Whelan and Chen Kong and Omkar Parkhi and Richard Newcombe and Yuheng Ren},
  title     = {Aria Digital Twin: A New Benchmark Dataset for Egocentric 3D Machine Perception},
  booktitle = {ICCV},
  pages     = {20133--20143},
  year      = {2023}
}

@article{zhang2023aptsim2real,
  author  = {Charles Y. Zhang and Ashish Shrivastava},
  title   = {{AptSim2Real}: Approximately-Paired Sim-to-Real Image Translation},
  journal = {arXiv preprint arXiv:2303.12704},
  year    = {2023}
}

@inproceedings{tobin2017domainrandomization,
  author    = {Josh Tobin and Rachel Fong and Alex Ray and Jonas Schneider and Wojciech Zaremba and Pieter Abbeel},
  title     = {Domain Randomization for Transferring Deep Neural Networks from Simulation to the Real World},
  booktitle = {IROS},
  year      = {2017}
}

@misc{zhou2025paibench,
  author        = {Fengzhe Zhou and Jiannan Huang and Jialuo Li and Deva Ramanan and Humphrey Shi},
  title         = {{PAI-Bench}: A Comprehensive Benchmark for Physical AI},
  year          = {2025},
  eprint        = {2512.01989},
  archiveprefix = {arXiv},
  primaryclass  = {cs.CV}
}

@article{nvidia2025cosmostransfer1,
  author  = {{NVIDIA}},
  title   = {{Cosmos-Transfer1}: Conditional World Generation with Adaptive Multimodal Control},
  journal = {arXiv preprint arXiv:2503.14492},
  year    = {2025}
}

@misc{nvidia2026cosmostransfer25,
  author       = {{NVIDIA}},
  title        = {{Cosmos-Transfer2.5}: Multi-Control World-to-World Generation},
  year         = {2026},
  howpublished = {\url{https://github.com/nvidia-cosmos/cosmos-transfer2.5}},
  note         = {Accessed September 15, 2026}
}

@misc{alibaba2025wan22fun,
  author       = {{Alibaba PAI Team}},
  title        = {{Wan2.2-Fun-Control}: Controllable Video Generation},
  year         = {2025},
  howpublished = {\url{https://github.com/aigc-apps/VideoX-Fun}},
  note         = {Accessed September 15, 2026}
}

@misc{minimax2026h3,
  author       = {{MiniMax}},
  title        = {{MiniMax H3}: An Open General-Purpose Multimodal Video Model},
  year         = {2026},
  howpublished = {\url{https://www.minimax.io/news/minimax-h3-open-source}},
  note         = {Accessed September 15, 2026}
}

@misc{bytedance2026seedance25,
  author       = {{ByteDance Seed Team}},
  title        = {{Seedance 2.5}: One-Take Creation with Flexible Referencing},
  year         = {2026},
  howpublished = {\url{https://seed.bytedance.com/en/blog/one-take-creation-flexible-referencing-introducing-seedance-2-5}},
  note         = {Accessed September 15, 2026}
}

@inproceedings{kerbl20233dgs,
  author    = {Bernhard Kerbl and Georgios Kopanas and Thomas Leimk{\"u}hler and George Drettakis},
  title     = {{3D Gaussian Splatting} for Real-Time Radiance Field Rendering},
  booktitle = {ACM SIGGRAPH},
  year      = {2023}
}

@inproceedings{wang2024dust3r,
  author    = {Shuzhe Wang and Vincent Leroy and Yohann Cabon and Boris Chidlovskii and J{\'e}r{\^o}me Revaud},
  title     = {{DUSt3R}: Geometric 3D Vision Made Easy},
  booktitle = {CVPR},
  year      = {2024}
}

@inproceedings{wang2025vggt,
  author    = {Jianyuan Wang and Minghao Chen and Nikita Karaev and Andrea Vedaldi and Christian Rupprecht and David Novotny},
  title     = {{VGGT}: Visual Geometry Grounded Transformer},
  booktitle = {CVPR},
  year      = {2025}
}

@misc{openai2026astra,
  author       = {{OpenAI}},
  title        = {{GPT-6 Astra}: A New Generation of Intelligence},
  year         = {2026},
  howpublished = {\url{https://openai.com/index/gpt-6-astra/}},
  note         = {Accessed September 15, 2026}
}
}

\end{document}